# Reproducibility Evaluation of SLANT Whole Brain Segmentation Across Clinical Magnetic Resonance Imaging Protocols

Yunxi Xiong [a], Yuankai Huo *[b], Jiachen Wang [a], L. Taylor Davis [c], Maureen McHugo [d], Bennett A. Landman [a,b,c]

[a] Computer Science, Vanderbilt University, Nashville, TN, USA 37235
[b] Electrical Engineering, Vanderbilt University, Nashville, TN, USA 37235
[c] Departments of Radiology and Radiological Sciences, Vanderbilt University Medical Center, Nashville, USA
[d] Department of Psychiatry and Behavioral Sciences, Vanderbilt University Medical Center, Nashville, TN, USA 37235
(* Corresponding Author, yuankai.huo@vanderbilt.edu)

**ABSTRACT**

Whole brain segmentation on structural magnetic resonance imaging (MRI) is essential for understanding neuroanatomical-functional relationships. Traditionally, multi-atlas segmentation has been regarded as the standard method for whole brain segmentation. In past few years, deep convolutional neural network (DCNN) segmentation methods have demonstrated their advantages in both accuracy and computational efficiency. Recently, we proposed the spatially localized atlas network tiles (SLANT) method, which is able to segment a 3D MRI brain scan into 132 anatomical regions. Commonly, DCNN segmentation methods yield inferior performance under external validations, especially when the testing patterns were not presented in the training cohorts. Recently, we obtained a clinically acquired, multi-sequence MRI brain cohort with 1480 clinically acquired, de-identified brain MRI scans on 395 patients using seven different MRI protocols. Moreover, each subject has at least two scans from different MRI protocols. Herein, we assess the SLANT method's intra- and inter-protocol reproducibility. SLANT achieved less than 0.05 coefficient of variation (CV) for intra-protocol experiments and less than 0.15 CV for inter-protocol experiments. The results show that the SLANT method achieved high intra- and inter- protocol reproducibility.

**Keywords**: SLANT, whole brain segmentation, MRI, coefficient of variation

## 1. INTRODUCTION

Historically, multi-atlas segmentation (MAS) [1-3] has been regarded as the de facto standard method for detailed whole brain segmentation on magnetic resonance image (MRI) (e.g., > 100 anatomical regions) due to its high accuracy. However, the heavy computational time (e.g. > 30 hours per scan) restricts the MAS whole brain segmentation on large-scale clinical cohorts. Recent advances in deep convolutional neural networks (DCNN) provide powerful tools to alleviate the computational time for whole brain segmentation [4-8], and even may result in higher accuracy compared with MAS such as the recently proposed spatially localized atlas network tiles (SLANT) method [9] (https://github.com/MASILab/SLANT_brain_seg). The SLANT method has been shown to achieve superior performance on small-scale validation methods as it was trained on >5000 MRI scans from >60 sites [10]. However, it has not been evaluated on large-scale clinically acquired cohorts as external validation. Figure 1 illustrates the quality of both intensity image and the segmentation has variations even if they are from the same subject within the same protocols. Therefore, it is essential to see if the segmentation method works for external cohorts other than the training data, which is also a major question to answer before applying the method in real clinical usage.

In this paper, we perform the external validation for the SLANT method on 1480 clinically acquired brain MRIs with seven different sequencing protocols. For each scan, 132 regions of interest (ROI) are identified by applying the SLANT method following the BrainCOLOR protocol [11]. Since it is difficult to obtain a large-scale external validation dataset with manual delineation, the reproducibility on ROI volumes of SLANT segmentation on the same subjects (but different scans) was used as the primary outcome metric. Specifically, the coefficient of variation (CV) was used as the metrics to evaluate the consistency of ROI volumes between the same clinical sequence protocols (intra-protocol), and across different clinical sequence protocols (inter-protocol). The SLANT segmentation method achieved, on average, a less than 0.05 coefficient of variation for intra-protocol from the same subject and a worst coefficient of variation 0.143 for inter-protocol sequence comparison.

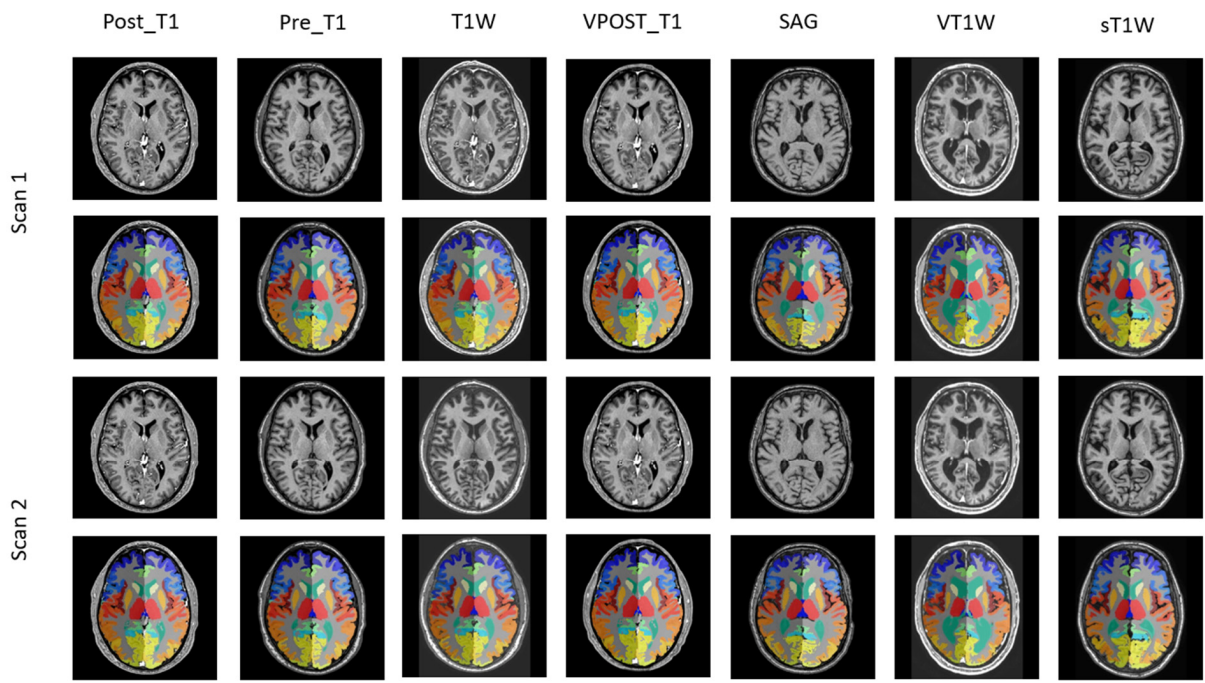

Figure 1. Seven protocols and both of intensity image and result segmentation of their scan-rescan is presented. The first four protocol sequences are from the same subject while the last three protocol sequences are from different subjects. "POST_T1" and "VPOST_T1" typically represents contrast-enhanced MRI T1 scans, while the "Pre_T1" typically indicates the T1 scans without contrast-enhancement. The "T1W" and "VT1W" could be a scan with contrast-enhancement, which is sometimes used in clinical scans. The protocol heterogeneity observed in this study highlights the need for methods that are robust to variations in clinical practice.

## 2. METHODS

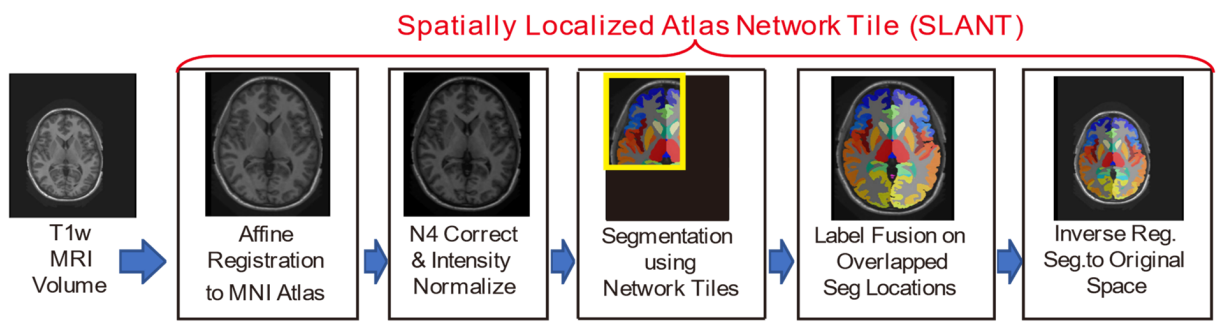

Figure 2. Workflow of the SLANT method. Whole brain segmentation method is presented, which combines canonical medical image processing methods (registration, harmonization, label fusion) with 3D network tiles.

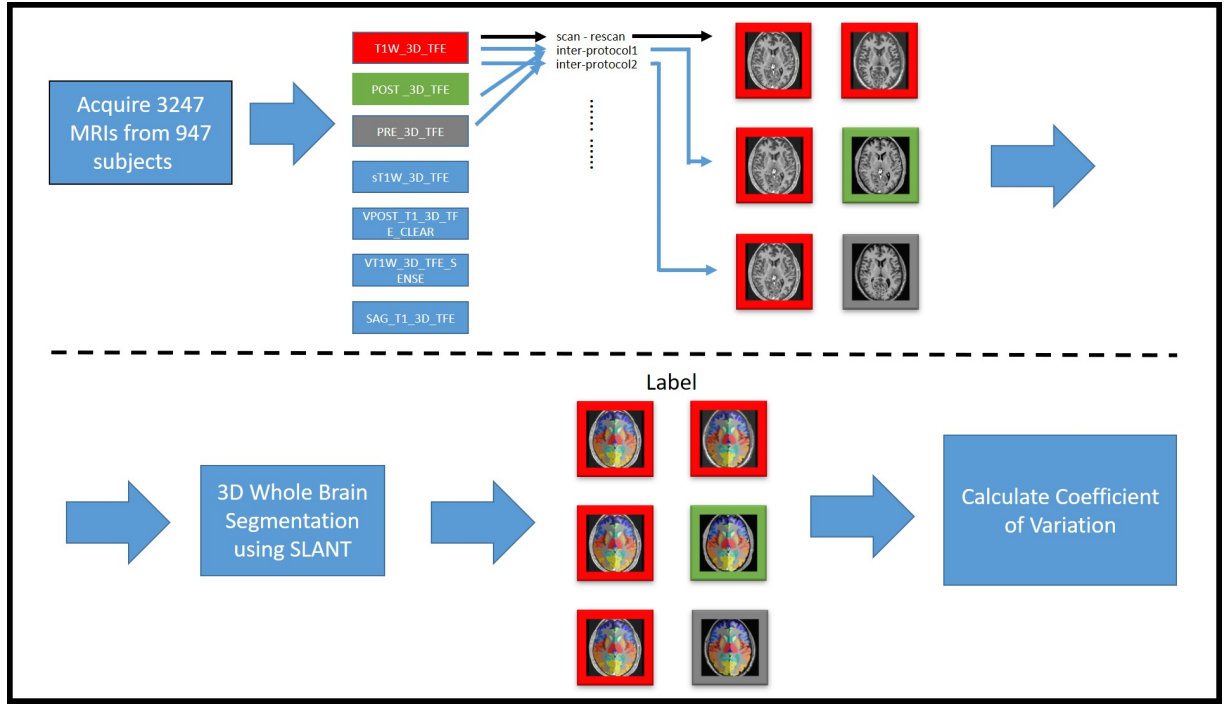

Figure 3. Workflow evaluating SLANT reproducibility across clinical protocol sequences. (1) 1840 T1 MRI scans were acquired to conduct quality assessment. (2) Seven different protocol sequences were selected where each protocol sequence has at least 30 MRIs. (3) SLANT segmentation was applied on all useable MRIs and generate labels and corresponding ROI volumes. (4) Coefficient of Variation (CV) was calculated between the same protocol sequences and across different protocol sequences.

**2.1 Quality assessment and protocol sequence selection**

The input image for the SLANT method was a single MRI T1 brain. All the data were retrieved in de-identified form under institutional review board approval from clinical scans studies conducted at the Vanderbilt University Medical Center (VUMC). For quality assurance (QA), the retrieved imaging data was visually inspected to verify that each scan has a complete brain without any missing tissue (field of view), without gross motion/imaging artifacts, and lacking large scale neuro-abnormality (e.g., visually apparent brain tumors). After QA, 1480 MRI scans from 395 unique patients from seven MRI sequence protocols were employed in the experiments, where each protocol has at least 30 MRIs to ensure we have sufficient data to perform statistical analyses.

**2.2 SLANT segmentation and ROI volume acquisition**

The proposed SLANT method is presented in Figure 2, which combines canonical medical image processing methods (registration, harmonization, label fusion) with 3D network tiles. 3D U-Net is used as each tile, whose deconvolutional channel numbers are modified to 133. The tiles are spatially overlapped in MNI space. All raw intensity images from seven selected protocols were used as input to run the SLANT segmentation (Figure 2). Since the SLANT method has been implemented as a Docker container [12], we used that Docker directly without any fine-tuning on the validation cohorts. The 1480 testing scans were analyzed for external validation for the SLANT segmentation. After segmentation, 132 regions of interest (ROI) were identified for each scan and each ROI's volume is calculated.

**2.3 ROI volume comparison between protocol sequences and coefficient of variation calculation**

With seven protocols, there are in total 28 possible paired protocol combinations where seven are within-sequence validations and the remaining 21 are cross-sequence validations. Note that we retrieved at least 30 subjects with MRI scans from following combinations: (1) SAG_T1_3D_TFE and VT1W_3D_TFE_SENSE, (2) SAG_T1_3D_TFE and

sT1W_3D_TFE, and (3) sT1W_3D_TFE and VT1W_3D_TFE_SENSE. Therefore, 25 combinations of protocol sequences were evaluated in this work.

Next, for each subject, we evaluate all MRI scans with a particular sequence protocol and then find protocol scans with the six other protocols. Then, we calculate the standard deviation and mean volumes of each of 132 ROI of the all combinations of sequence protocols for the same subjects. After calculating the standard deviation and mean volumes of one combination for each ROI, we calculate the coefficient of variation (CV) of each ROI using following equation:

$$CV = \frac{2\theta(S_1 - S_2)}{\mu(S_1 + S_2)} \quad (1)$$

where $S_1$ indicates an array of one ROI's volume from all subjects of one particular protocol sequence and $S_2$ indicates an array of one ROI's volume from all subjects of another protocol sequence with each element in $S_1$ and its corresponding element in $S_2$ from the same subject but in different scans. $\theta$ indicates the standard deviation and $\mu$ denotes the mean.

## 3. EXPERIMENTS

### 3.1 Data

Table 1 The number of subjects for 28 validation combinations

|       | T1  | POST | PRE | sT1W | VPOST | VT1W | SAG |
|-------|-----|------|-----|------|-------|------|-----|
| T1    | 619 | 434  | 137 | 50   | 57    | 154  | 28  |
| POST  |     | 177  | 384 | 7    | 60    | 31   | 14  |
| PRE   |     |      | 223 | 8    | 73    | 43   | 18  |
| sT1W  |     |      |     | 43   | 2     | 0    | 42  |
| VPOST |     |      |     |      | 35    | 7    | 0   |
| VT1W  |     |      |     |      |       | 49   | 0   |
| SAG   |     |      |     |      |       |      | 39  |

*1195 T1 MRIs volumes from dataset provided by Vanderbilt University Medical Center were used to evaluate the reproducibility of the SLANT method with the number of combination between each protocol sequence.

Data eligible for the study were identified from the Research Derivative, a database of clinical and related data derived from the Vanderbilt University Medical Center's (VUMC) clinical systems and restructured for research. The Research Derivative was searched for individuals between the ages of 18 and 80 years old who had a T1-weighted structural MRI with <=1.5mm voxel resolution and sodium lab values obtained between 2016-2017. The T1 images for each patient were then retrieved in deidentified form from ImageVU, a research database of magnetic resonance and computed tomography images that are obtained in the clinical setting at VUMC.

Seven sequence types were used in the comparisons distinguished by the names of sequences. For naming, T1_3D_TFE ("T1") means T1 weighted 3D turbo field echo (TFE) MRI at 1mm slice thickness from Philips' nomenclature. "POST" and "PRE" means the TFE MRI acquired after and before gadolinium administration respectively. "SAG" means the TFE MRI acquired from sagittal acquisition. 1480 T1 MRIs volumes with seven sequencing protocols acquired from Vanderbilt University Medical Center were used to evaluate the reproducibility of the SLANT method with the number of combinations between each protocol sequence (Table 1). In Table 1, some cells are intentionally left blank because they are the same comparison if the order is reversed (e.g., T1 vs. POST has the same number of comparisons as POST vs. T1).

### 3.2 Experiments

The total number of MRI scans used in the within-protocol experiments equals the total number of comparisons between the same protocol sequence across the diagonal numbers in Table 1(619 + 177 + 223 + 43 + 35 + 49 + 49 = 1195). The numbers of inter protocol subjects are presented in the remaining elements in Table 1. Note that there were more inter-protocol comparisons than intra-protocol comparisons because every MRI of one particular protocol was used to compare with all MRIs of another protocol sequence of the same subjects. In general, the grand total number of comparison between all combinations of protocol sequences is 2734.

The SLANT pretrained model in the SLANT Docker (https://github.com/MASILab/SLANT_brain_seg) was used directly to segment all input scans with a variety of input resolutions and voxel sizes. For the experiments in this study, the SLANT segmentation Docker was run on NVIDIA GeForce 1080 TI GPU with 11GB memory. The SLANT Docker takes approximately 15 minutes to segment a single input MRI scan.

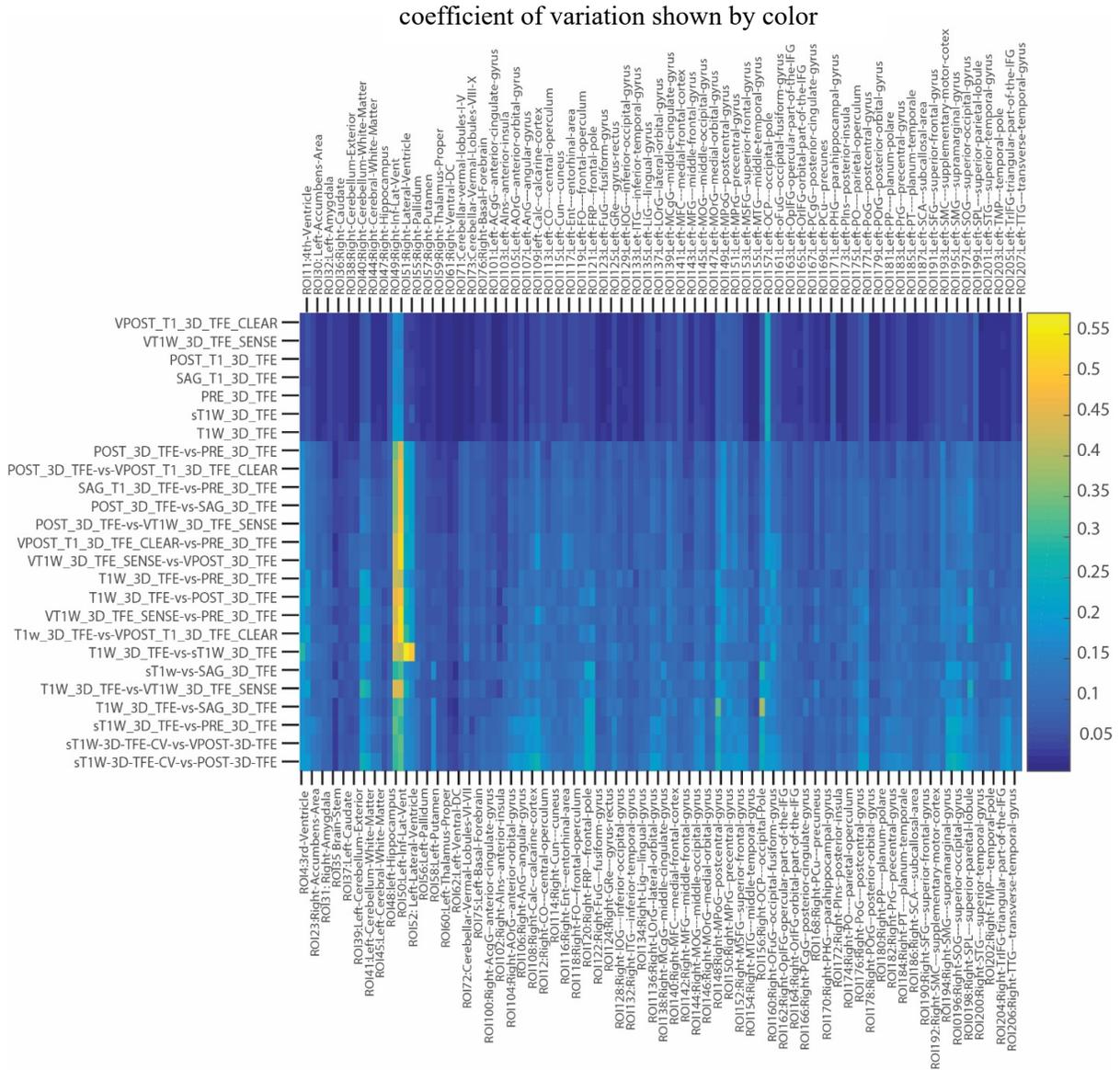

Figure 3. The reproducibility of SLANT method was shown. The CV of variation of each ROI of each comparison between protocol sequences were demonstrated. The color of each label corresponds to the CV of variation value shown in the color bar.

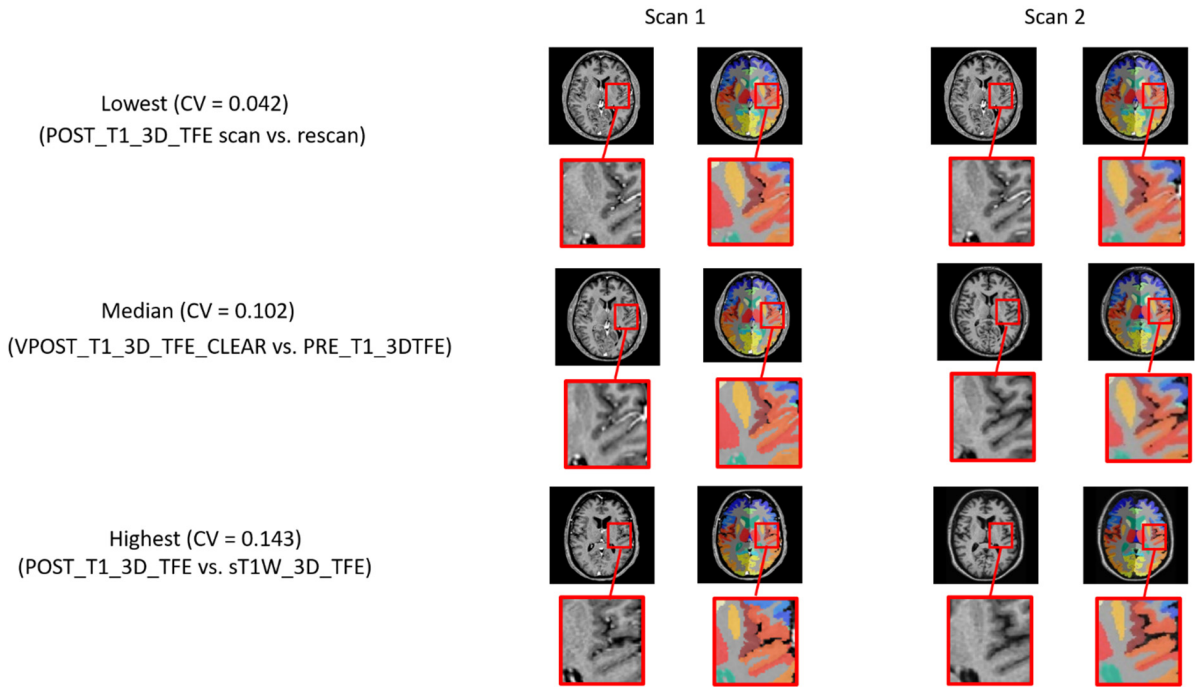

Figure 4. Qualitative reproducibility of the SLANT method. In the lowest CV case, the segmentations between two scans within the same protocols look highly similar, while in the highest CV case the segmentations have more inter-scan differences.

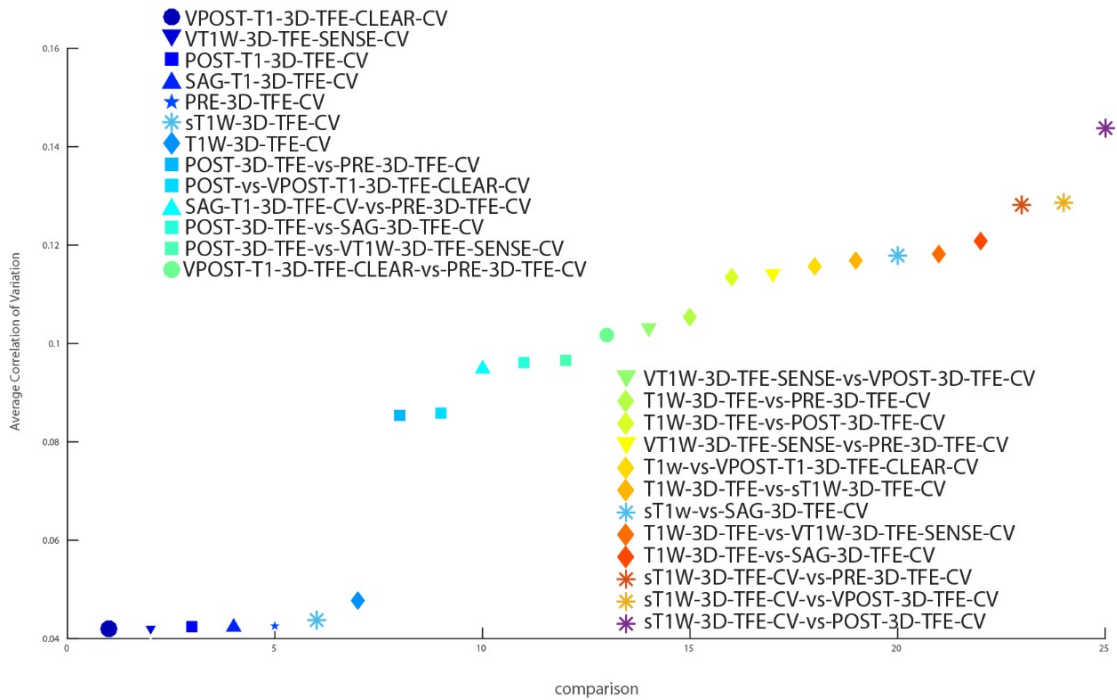

Figure 5. The average CV between each protocol sequence comparison. Each shape in the figure represents the comparison with one particular protocol sequence.

## 4. RESULTS

We show the CV of each combination of protocol sequences using quantitative (Figures 4 and 5) as well as qualitative results (Figure 6). In Figure 4, every cell in the matrix represents the CV for one ROI from one comparison of protocol sequences and this matrix is sorted by the average CV. In particular, we note that colors on the top seven rows were much darker than the rest of the matrix, which shows the CV of variation is much lower for this set of sequences compared to the others. Those seven rows match seven comparisons between the same protocols, which is expected because the CV between the same protocol sequence is supposed to be much lower than that between inter-protocol comparisons. In Figure 5, we show the average CV of all combinations of protocol sequences. Each shape represents a particular protocol sequence compared with other protocol sequences. We find similar results to those shown in Figure 4; the average CV between the same protocol is much lower than that of inter-protocol comparisons. as the seven points on the bottom of the figure corresponds to the seven intra-protocol comparisons. However, the CV scores of inter-protocol comparisons are still reasonable, where the worst average CV between two protocol sequences is around 0.143. In Figure 6, three groups of comparisons were presented, which presents the qualitative examples of highest, median, and lowest CV using SLANT. With a worst correlation variation 0.143 between protocol sequence POST_T1_3D_TFE and protocol sequence sT1W_3D_TFE and a best correlation of variation of only 0.042 in intra-protocol POST_T1_3D_TFE comparison.

## 5. CONCLUSION AND DISCUSSION

In this study, we evaluate the reproducibility of SLANT method by means of calculating the coefficient of variation between each combination of clinical protocol sequences. Based on Figure 4, the CV is lower in intra-protocol sequence comparisons than in inter-protocol sequence comparisons, which is expected as the way of acquiring the data is slightly different if protocol sequences were different. In Figure 5, the result showed that the SLANT method achieved a CV that is less than 0.05 for all intra-protocol comparisons. Considering the inter-protocol comparisons as well, the lowest CV is 0.042 (comparison between the same protocol sequences), while the highest CV is 0.143 (comparison between the different protocol sequences). We conclude that SLANT method yields a high reproducibility across clinical protocol sequences.

There are also several limitations of this work. Based on Figure 4, we see two yellow strips in inter-protocol comparison area and light blue in intra-protocol comparison area, which means these two ROIs (49 Right-Inf-Lat-Vent and 50 Left-Inf-Lat-Vent) have notably and consistently higher CV across any protocol sequence comparison. This phenomenon requires further investigation. Another limitation is that we only evaluated the SLANT whole brain segmentation volumetry in this study. In the future, we would like to perform the external validation on other representative whole brain segmentation methods and include overlap measures, surface distances, and comparisons with expert human raters.

## 6. ACKNOWLEDGEMENT

This research was supported by NSF CAREER 1452485, NIH grants 5R21EY024036, 1R21NS064534, 2R01EB006136 (Dawant), 1R01EB017230 (Landman), 1R03EB012461 (Landman) and R01NS095291 (Dawant). InCyte Corporation (Abramson/Landman). This research was conducted with the support from and the Charlotte and Donald Test Fund, Intramural Research Program, National Institute on Aging, NIH. This study was in part using the resources of the Advanced Computing Center for Research and Education (ACCRE) at Vanderbilt University, Nashville, TN. This project was supported in part by ViSE/VICTR VR3029 and the National Center for Research Resources, Grant UL1 RR024975-01, and is now at the National Center for Advancing Translational Sciences, Grant 2 UL1 TR000445-06. We gratefully acknowledge the support of NVIDIA Corporation with the donation of the Titan X Pascal GPU used for this research. The imaging dataset(s) used for the analysis described were obtained from ImageVU, a research resource supported by the VICTR CTSA award (ULTR000445 from NCATS/NIH), Vanderbilt University Medical Center institutional funding and Patient-Centered Outcomes Research Institute (PCORI; contract CDRN-1306-04869).

# 7. REFERENCES

[1] A. J. Asman, A. S. Dagley, and B. A. Landman, "Statistical label fusion with hierarchical performance models," Proc Soc Photo Opt Instrum Eng, 9034, 90341E (2014).

[2] A. J. Asman, Y. Huo, A. J. Plassard *et al.*, "Multi-atlas learner fusion: An efficient segmentation approach for large-scale data," Medical image analysis, 26(1), 82-91 (2015).

[3] Y. Huo, A. J. Asman, A. J. Plassard *et al.*, "Simultaneous total intracranial volume and posterior fossa volume estimation using multi‐atlas label fusion," Human brain mapping, 38(2), 599-616 (2017).

[4] A. de Brébisson, and G. Montana, "Deep neural networks for anatomical brain segmentation," arXiv preprint arXiv:1502.02445, (2015).

[5] R. Mehta, A. Majumdar, and J. Sivaswamy, "BrainSegNet: a convolutional neural network architecture for automated segmentation of human brain structures," Journal of Medical Imaging, 4(2), 024003 (2017).

[6] C. Wachinger, M. Reuter, and T. Klein, "DeepNAT: Deep convolutional neural network for segmenting neuroanatomy," NeuroImage, (2017).

[7] W. Li, G. Wang, L. Fidon *et al.*, "On the compactness, efficiency, and representation of 3D convolutional networks: brain parcellation as a pretext task." 348-360.

[8] A. G. Roy, S. Conjeti, D. Sheet *et al.*, "Error Corrective Boosting for Learning Fully Convolutional Networks with Limited Data." 231-239.

[9] Y. Huo, Z. Xu, K. Aboud *et al.*, "Spatially Localized Atlas Network Tiles Enables 3D Whole Brain Segmentation from Limited Data," arXiv preprint arXiv:1806.00546, (2018).

[10] Y. Huo, K. Aboud, H. Kang *et al.*, "Mapping lifetime brain volumetry with covariate-adjusted restricted cubic spline regression from cross-sectional multi-site MRI." 81-88.

[11] A. Klein, T. Dal Canton, S. S. Ghosh *et al.*, "Open labels: online feedback for a public resource of manually labeled brain images," 16th Annual Meeting for the Organization of Human Brain Mapping. (2010).

[12] D. Merkel, "Docker: lightweight linux containers for consistent development and deployment," Linux Journal, 2014(239), 2 (2014).